\documentclass[pdflatex,sn-basic,iicol]{sn-jnl}

\usepackage{graphicx}%
\usepackage{multirow}%
\usepackage{amsmath,amssymb,amsfonts}%
\usepackage{amsthm}%
\usepackage{mathrsfs}%
\usepackage[title]{appendix}%
\usepackage{xcolor}%
\usepackage{textcomp}%
\usepackage{manyfoot}%
\usepackage{booktabs}%
\usepackage{algorithm}%
\usepackage{algorithmicx}%
\usepackage{algpseudocode}%
\usepackage{listings}%

\usepackage[utf8]{inputenc}
\usepackage[T1]{fontenc}

\theoremstyle{thmstyleone}%
\theoremstyle{thmstyletwo}%

\theoremstyle{thmstylethree}%

\begin{document}

\title[Article Title]{A Study of ASR Adaptation and Representation Dimensionality Reduction in Persian Speech Emotion Recognition Using Whisper}


\author[1]{\fnm{Ali} \sur{Shendabadi}}\email{alishendabadi@ut.ac.ir}

\author[1]{\fnm{Parnia} \sur{Izadirad}}\email{parniaizadirad@ut.ac.ir}

\author*[1]{\fnm{Mostafa} \sur{Salehi}}\email{mostafa\_salehi@ut.ac.ir}

\affil*[1]{\orgdiv{Department of Data Science and Technology}, \orgdiv{School of Intelligent Systems Engineering}, \orgdiv{College of Interdisciplinary Sciences and Technologies}, \orgname{University of Tehran}, \orgaddress{\street{North Kargar}, \city{Tehran}, \postcode{1417935840}, \country{Iran}}}


\abstract{Speech Emotion Recognition (SER) in low-resource languages remains a challenging problem due to limited labeled data. In this work, we study the use of Whisper for Persian SER with a particular focus on representation dimensionality reduction and language-specific model adaptation. We propose a SER framework in which frame-level embeddings extracted from the Whisper encoder are reduced in dimensionality using PCA, eliminating the need for learned projection layers and substantially reducing the number of trainable parameters. The reduced representations are aggregated using an attention-based pooling mechanism and classified with a lightweight prediction head. In addition, we investigate whether fine-tuning Whisper on a Persian automatic speech recognition (ASR) task improves downstream SER performance. Experiments conducted on the ShEMO dataset under a speaker-independent evaluation protocol show that PCA-based dimensionality reduction consistently improves emotion recognition performance while reducing training latency and memory usage. ASR fine-tuning yields only modest gains for SER, suggesting limited transfer from language adaptation to emotion-related representations under the evaluated conditions. These findings provide practical insights into the efficient use of large pretrained speech models for emotion recognition in low-resource languages.}




\maketitle

\section{Introduction}

\begin{figure}[h!]
\centering
\includegraphics[scale=0.1]{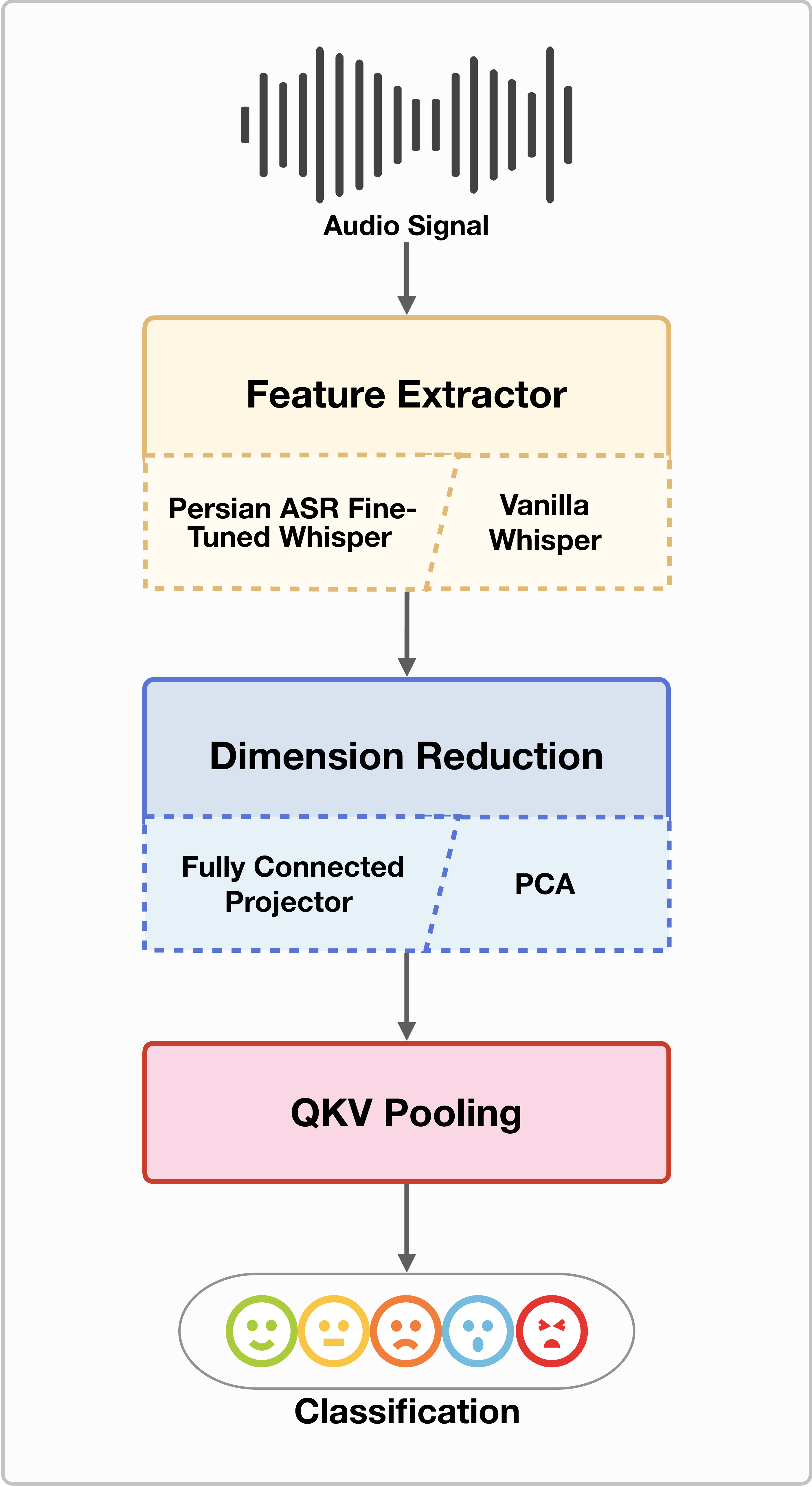}
\label{fig:g_abstract}
\caption{In this framework frame-level representations are extracted using the encoder of a pretrained Whisper model. Dimensionality is reduced utilizing either PCA or a fully connected projection layer. The resulting embeddings are aggregated into a fixed-length utterance-level representation using an attention-based QKV pooling mechanism. A lightweight classification head is then applied to map the utterance-level embedding to emotion categories. Additionally, the effect of adapting Whisper to Persian through ASR fine-tuning is investigated prior to SER training.}
\end{figure}

SER has become an important research topic within the broader field of speech processing, driven by its potential to enable more natural and effective human--machine interaction. By analyzing emotional cues embedded in speech signals, SER systems can enhance a wide range of applications, including voice assistants \cite{voice_assistant}, intelligent tutoring systems \cite{tutoring}, and customer support analytics \cite{customer_support1, customer_support2}. In healthcare and assistive settings, SER has shown promise for supporting the assessment of psychological states and improving accessibility for individuals with physical or cognitive impairments \cite{health, health2, forensics_health}. In addition, SER is increasingly relevant in forensic and security-related contexts \cite{forensics, forensics_health}, such as analyzing emotional states in investigative interviews, emergency calls, and judicial proceedings, where understanding affective cues may provide complementary information for decision-making \cite{dstm}.

Despite notable progress enabled by deep learning, robust emotion recognition from speech remains a challenging problem. The task is particularly difficult for low-resource languages such as Persian, where the availability of emotion-labeled speech data is limited and linguistic variability is high. These constraints make it difficult to train expressive models without overfitting. They also hinder the direct transfer of approaches developed primarily for high-resource languages.

Early SER systems relied largely on handcrafted acoustic features, such as prosodic, spectral, and voice quality descriptors. While these features capture important paralinguistic cues related to emotion, they may not fully reflect higher-level information conveyed through speech. Emotional expression is influenced not only by acoustic patterns such as pitch, energy, and timing, but also by linguistic and semantic content. Motivated by this observation, recent studies have explored multimodal SER frameworks that combine audio and textual information, often demonstrating improved performance when semantic cues are explicitly incorporated \cite{multimodal}. More recently, the field has shifted toward the use of large pretrained speech models as feature extractors for SER. Models such as Wav2Vec 2.0 \cite{wav2vec}, HuBERT \cite{hubert}, and Whisper \cite{whisper} learn rich representations from large-scale speech corpora and have achieved state-of-the-art performance across a variety of downstream tasks. Compared to traditional handcrafted features, these representations capture more abstract and informative characteristics of speech.

However, the use of such pretrained models introduces two key challenges in low-resource SER settings. First, the extracted frame-level representations are typically high-dimensional---for example, Whisper-small produces 768-dimensional embeddings---leading to a large number of trainable parameters when followed by a trainable projection layer. This often results in overfitting and inefficient training when emotion-labeled data are scarce \cite{shendabadi}. Second, although many pretrained speech models are multilingual, low-resource languages such as Persian are underrepresented in their pretraining data, potentially limiting the language-specific information encoded in their representations. For instance in whisper's pretraining data there were just 24 hours of Persian ASR speech and 392 hours of Persian to English Speech translation data.

In this work, we address these challenges by proposing an efficient SER framework for Persian that builds upon the Whisper model. To mitigate overfitting caused by high-dimensional representations, we employ principal component analysis (PCA) as an unsupervised dimensionality reduction technique, replacing trainable projection layer and eliminating additional trainable parameters. Furthermore, motivated by the natural expectation that language adaptation may improve downstream performance, we investigate the effect of fine-tuning Whisper on a Persian ASR task prior to SER training. This allows us to systematically examine whether improved language-specific representations translate into gains for emotion recognition. Experiments are conducted on the ShEMO \cite{shemo} dataset under a speaker-independent evaluation protocol.

The main contributions of this work are summarized as follows:
\begin{itemize}
\item We demonstrate that unsupervised dimensionality reduction using PCA can effectively replace learned fully connected projection layers, leading to improved SER performance as well as substantial reductions in training latency and memory usage.

\item We systematically investigate the impact of language-specific ASR fine-tuning of Whisper on downstream SER performance for Persian and show that, under the evaluated conditions, the resulting gains are modest.
\end{itemize}

\section{Related Works}
In recent years, SER has increasingly relied on large pretrained speech models as feature extractors, replacing traditional handcrafted acoustic representations. Among these models, Whisper has attracted growing attention due to its strong multilingual capabilities and robust performance across a wide range of downstream speech tasks.

Several studies have explored the effectiveness of Whisper representations for SER. In \cite{b1}, representations extracted from Whisper-large are combined with handcrafted acoustic features such as MFCCs, leading to performance improvements of approximately 1\% on EMODB and 0.1\% on RAVDESS. A broader comparison of pretrained speech models is presented in \cite{b3}, where multiple models are evaluated on Spanish SER across several datasets. Of particular relevance to our work, this study compares the original Whisper large-v2 model with a Spanish ASR-fine-tuned variant, reporting only limited or no gains from language-specific fine-tuning---an observation we replicate and further examine in the context of Persian SER.

Whisper has also been evaluated for SER in other low-resource language settings. In \cite{b4}, the KEDAS dataset is used to assess Whisper's performance for Arabic emotion recognition. The authors report relatively low accuracy (approximately 37\% across five emotion classes) and conclude that task-specific fine-tuning of Whisper is likely necessary to achieve competitive results. A different approach is proposed in \cite{b5}, where Whisper-large is directly fine-tuned for SER on the IEMOCAP dataset and subsequently distilled into a smaller Whisper encoder. This framework achieves a 7.21$\times$ reduction in model size while retaining 99.99\% of the teacher model's unweighted average recall, reaching 79.82\% weighted and 81.32\% unweighted average recall.

Beyond purely acoustic representations, some works have explored the use of Whisper's decoder outputs to incorporate textual information into SER. For example, \cite{b7} utilizes both the encoder and decoder components of Whisper to perform bimodal emotion recognition, demonstrating that leveraging linguistic content alongside acoustic cues can yield notable performance improvements. In a related direction, \cite{b8} investigates SER for Mongolian, another low-resource language, by extracting representations from different Whisper encoder layers and aggregating them using an attentive fusion mechanism. Their findings highlight the importance of layer selection and pooling strategies for emotion recognition in underrepresented languages.

Alternative classification approaches based on Whisper features have also been proposed. In \cite{b9}, Whisper-based representations extracted from the IEMOCAP dataset are classified using a Kolmogorov--Arnold network, demonstrating that non-standard classifiers can be effectively combined with pretrained speech embeddings. Additionally, \cite{b6} provides an in-depth analysis of the internal mechanisms of Whisper adaptation for SER, focusing on the effects of LoRA-based fine-tuning and offering insights into how parameter-efficient adaptation influences emotion-related representations.

Research on Persian SER using Whisper remains relatively limited. A recent study in \cite{b2} employs Whisper to automatically transcribe the ShEMO dataset and then vectorizes the resulting text using FastText embeddings. These textual features are combined with handcrafted acoustic descriptors such as MFCCs and zero-crossing rate, with a differential evolution algorithm applied for feature selection. While innovative, this framework does not exploit Whisper's encoder or decoder embeddings directly, despite the substantial computational cost of ASR inference, instead relying on classical text representations.

in \cite{shendabadi} two attention-based pooling techniques---Multi-head Attentive Average Pooling and QKV Pooling---are utilized to reduce the dimensionality of Whisper's representations while retaining emotional information. The authors evaluated their method on English (IEMOCAP) and Persian (ShEMO) datasets, employing Whisper's Tiny and Small variants. Their multi-head QKV architecture achieved state-of-the-art results on the ShEMO dataset, yielding a 2.47\% improvement in unweighted accuracy. Additionally, they compared different Whisper encoder layers and observed that intermediate layers performed better for SER on the Persian dataset.

Together, these studies underscore both the growing interest in applying Whisper to SER and the persistent challenges associated with low-resource languages such as Persian. They also highlight open questions regarding the role of dimensionality reduction, pooling strategies, and language-specific adaptation---issues that are directly addressed in the present work.

\section{Methodology}
\subsection{Overview of the Proposed SER Framework}

Figure \ref{fig:g_abstract} illustrates the overall architecture of the proposed Persian SER framework. Given an input utterance, frame-level representations are first extracted using the encoder of a pretrained Whisper model. Due to the high dimensionality of these representations and the limited size of available Persian emotion data, PCA is applied to reduce dimensionality in an unsupervised manner. The resulting frame-level embeddings are then aggregated into a fixed-length utterance-level representation using an attention-based query--key--value (QKV) pooling mechanism \cite{shendabadi}. Finally, a lightweight classification head maps the utterance-level embedding to emotion categories. In addition to using the original Whisper model, we investigate the effect of adapting Whisper to Persian via ASR fine-tuning prior to SER training.

Formally, let $\mathbf{x}$ denote the input waveform. The whole pipeline can be summarized as
\begin{subequations}
\label{eq:pipeline}
\begin{align}
H &= \text{WhisperEnc}(\mathbf{x}), \label{eq:pipe_whisper}\\
Z &= \text{PCA}(H), \label{eq:pipe_pca}\\
\mathbf{a} &= \text{QKVPool}(Z), \label{eq:pipe_qkv}\\
\hat{\mathbf{y}} &= \text{Softmax}\big(\text{FC}(\mathbf{a})\big), \label{eq:pipe_out}
\end{align}
\end{subequations}
where $\hat{\mathbf{y}}\in\mathbb{R}^{C}$ is the predicted distribution over the $C$ emotion classes. Throughout this section, $T=1500$ denotes the number of frames per utterance, $H\in\mathbb{R}^{T\times 768}$ the Whisper encoder output, $Z\in\mathbb{R}^{T\times 512}$ the PCA-reduced sequence, and $\mathbf{a}\in\mathbb{R}^{512}$ the pooled utterance-level vector.

\subsection{Emotion Representation Extraction from Whisper}
Whisper is a transformer-based sequence-to-sequence speech model trained on large-scale multilingual data. In this work, we employ the Whisper-small variant due to its favorable balance between representational capacity and computational efficiency. Given an input speech signal, Whisper's encoder produces a sequence of frame-level hidden representations at each transformer layer, where each frame is represented by a 768-dimensional embedding and there are 1500 embedding vectors for a speech input file.

Different encoder layers capture complementary information ranging from low-level acoustic cues to higher-level linguistic and semantic features. We extract frame-level representations from the last whisper encoder layer and use them as input features for emotion recognition. The Whisper encoder parameters are kept frozen during SER training to mitigate overfitting in the low-resource setting.

Let $X\in\mathbb{R}^{80\times 3000}$ be the log-Mel spectrogram of the utterance (padded or truncated to 30 seconds) computed by the Whisper processor. The frame-level representation from the last encoder layer $L$ is
\begin{equation}
\begin{split}
H &= \text{WhisperEnc}^{(L)}(X)\\
  &= [\mathbf{h}_1,\dots,\mathbf{h}_T]^\top \in \mathbb{R}^{T\times 768}.
\end{split}
\label{eq:whisper}
\end{equation}

\subsection{PCA-Based Dimensionality Reduction}
The high dimensionality of Whisper encoder representations poses a significant challenge for SER on small datasets such as Shemo. In previous works, dimensionality reduction was performed using a learned fully connected projection layer that mapped 768-dimensional frame-level representations to 256 dimensions \cite{shendabadi}. For Whisper-small, this projection introduced $768\times256=196{,}608$ trainable parameters, which is substantial relative to the size of the available emotion-labeled data (approximately 3,000 utterances). This often led to early overfitting and limited the ability to increase model complexity in other components.

\begin{figure*}[h!]
\centering
\includegraphics[scale=0.7]{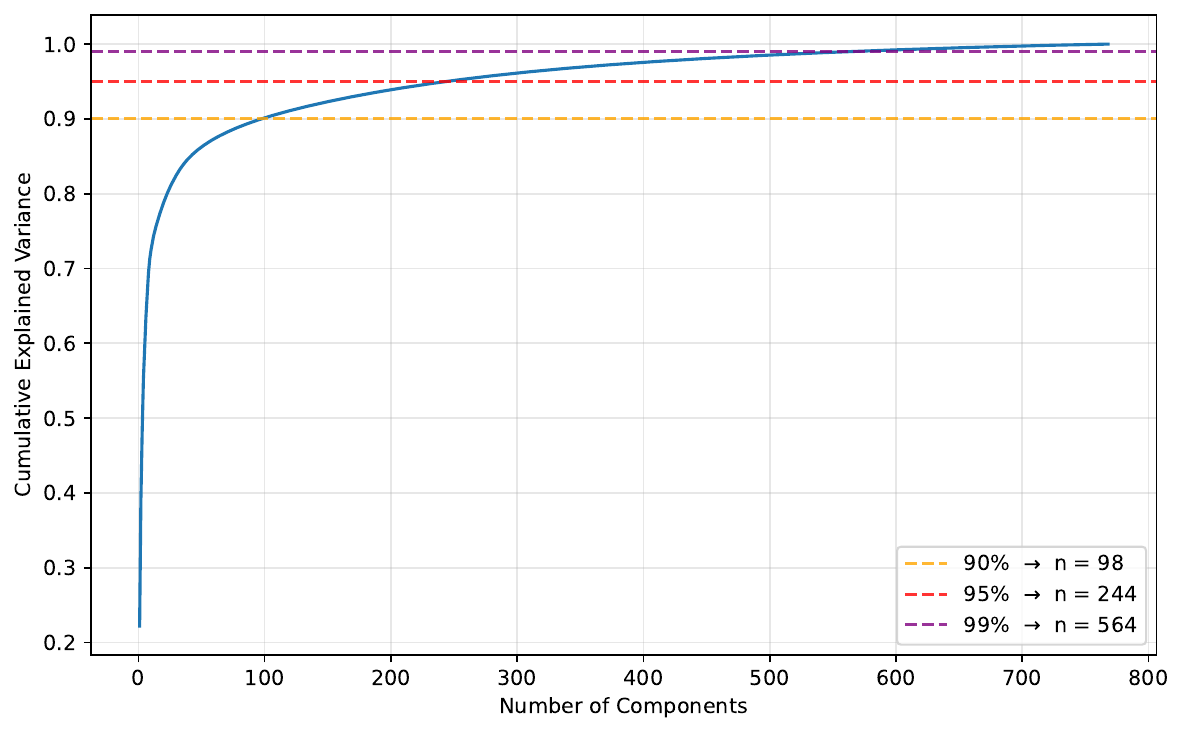}
\caption{Varience retention analysis scree plot for PCA}
\label{fig:scree}
\end{figure*}

To address this issue, we replace the learned projection layer with PCA, an unsupervised dimensionality reduction technique that introduces no additional trainable parameters. PCA is fitted using frame-level representations extracted from the training split only, and the learned transformation is subsequently applied to validation and test data, ensuring that no information from evaluation sets is used during model training.

Each frame is then projected onto the first $k=512$ principal components, whose mean $\boldsymbol{\mu}_{\text{PCA}}$ and projection matrix $W_{\text{PCA}}\in\mathbb{R}^{768\times 512}$ are estimated from training-split frames only and kept fixed afterwards:
\begin{equation}
\begin{split}
\mathbf{z}_t &= W_{\text{PCA}}^{\top}\big(\mathbf{h}_t-\boldsymbol{\mu}_{\text{PCA}}\big),\\
Z &= [\mathbf{z}_1,\dots,\mathbf{z}_T]^{\top}\in\mathbb{R}^{T\times 512}.
\end{split}
\label{eq:pca_proj}
\end{equation}
This transformation adds no trainable parameters.

To determine an appropriate target dimensionality, we conduct a variance retention analysis on Whisper representations. This analysis is used solely for exploratory purposes to assess the intrinsic dimensionality of the feature space. Results in figure \ref{fig:scree} indicate that retaining approximately 95\% of the total variance requires at least 244 principal components and retaining 99\% of it requires 564 principal components. Based on this observation we reduce the dimensionality of frame-level representations to 512.

\subsection{Attention-Based Pooling}
After dimensionality reduction, the sequence of frame-level embeddings is aggregated into a single utterance-level representation. We employ QKV attention-based pooling mechanism, which is introduced in CLIP paper \cite{clip} and also utilized in our previous work for the first time in SER domain to the best of our knowledge. This pooling method learns to assign higher weights to frames that are more informative for emotion recognition, enabling the model to focus on emotionally salient regions of an utterance.

Following \cite{shendabadi}, the query is conditioned on the global average of the frame embeddings, while keys and values are computed from all frames:
\begin{subequations}
\label{eq:qkv}
\begin{align}
\boldsymbol{\mu}(Z) &= \frac{1}{T}\sum_{t=1}^{T}\mathbf{z}_t, \\
\text{Head}_i &= \text{Softmax}\!\left(\frac{\big(\boldsymbol{\mu}(Z)W_i^{Q}\big)\big(ZW_i^{K}\big)^{\top}}{\sqrt{d_k}}\right)ZW_i^{V}, \\
\mathbf{a} &= \text{Concat}(\text{Head}_1,\dots,\text{Head}_h)\,W^{O},
\end{align}
\end{subequations}
where $h$ is the number of heads, $W_i^{Q},W_i^{K}\in\mathbb{R}^{512\times d_k}$, $W_i^{V}\in\mathbb{R}^{512\times d_v}$, and $W^{O}\in\mathbb{R}^{(h\,d_v)\times 512}$, so that $\mathbf{a}\in\mathbb{R}^{512}$. The softmax is computed over the $T$ frames, so the resulting attention weights express the relative importance of each frame for the utterance-level representation.

\subsection{Whisper Fine-Tuning for Persian ASR}
Although Whisper demonstrates strong zero-shot performance across multiple languages, Persian remains underrepresented in its pretraining data. Moreover, emotion recognition benefits not only from acoustic cues but also naturally from linguistically grounded representations. We therefore hypothesize that adapting Whisper to Persian through ASR fine-tuning can improve the suitability of its encoder representations for downstream Persian SER due to its improved knowledge about Persian language and speech in the process of ASR finetuning.

To this end, we fine-tune Whisper-small on a Persian ASR dataset collected from YouTube \cite{pourmand2024asr}, consisting of approximately 340 hours of speech. Parameters in all the encoder layers and 10 of decoder layers are updated using the standard sequence-to-sequence cross-entropy objective
\begin{equation}
\mathcal{L}_{ASR}=-\sum_{j=1}^{J}\log p_{\theta}\big(y_j\mid y_{<j},X\big),
\label{eq:asr_loss}
\end{equation}
where $y_1,\dots,y_J$ are the reference transcript tokens and $\theta$ denotes the updated Whisper parameters. ASR performance is evaluated using word error rate (WER). Fine-tuning reduces WER from 45\% to 37\%, indicating effective adaptation to Persian speech.

Following ASR fine-tuning, the adapted Whisper encoder is used as a feature extractor for SER, while the remainder of the SER pipeline remains unchanged. This allows for a controlled comparison between representations obtained from the original and ASR-adapted Whisper models.

\subsection{Classification Head and Training Strategy}
The utterance-level embedding produced by the pooling module is passed to a lightweight classification head consisting of a single fully connected layer with softmax activation:
\begin{equation}
\hat{\mathbf{y}}=\text{Softmax}\big(W_c\,\mathbf{a}+\mathbf{b}_c\big),\quad W_c\in\mathbb{R}^{C\times 512},\ \mathbf{b}_c\in\mathbb{R}^{C}.
\label{eq:cls}
\end{equation}
The model is trained using cross-entropy loss over a mini-batch of $B$ samples:
\begin{equation}
\mathcal{L}_{CE}=-\frac{1}{B}\sum_{n=1}^{B}\sum_{c=1}^{C}y_{n,c}\log\hat{y}_{n,c},
\label{eq:ce}
\end{equation}
where $y_{n,c}$ is the one-hot ground-truth label. During SER training, Whisper encoder parameters remain frozen and only the pooling and classification components, i.e., $\{W_i^{Q},W_i^{K},W_i^{V}\}_{i=1}^{h}$, $W^{O}$, $W_c$, and $\mathbf{b}_c$, are optimized. This design choice is motivated by the limited size of the emotion-labeled dataset and aims to reduce overfitting.
\section{Experimental Setup}
\subsection{Data}
ShEMO \cite{shemo} is a Persian emotional speech corpus comprising 3,000 utterances drawn from roughly 3.5 hours of online radio drama and recorded by 87 native speakers. Each sample is annotated with one of six emotion categories: anger, fear, happiness, sadness, surprise, or neutral. The dataset exhibits a strong class imbalance, with anger and neutral together accounting for about 70\% of the data, sadness contributing approximately 15\%, and happiness and surprise forming a relatively small portion. Owing to its limited representation, the fear class is excluded in line with prior work. Following established practice in earlier studies and to facilitate fair comparability of results, the data are split into 10 speaker-independent folds.
\subsection{Training and Evaluation Details}
Most hyperparameters are selected in accordance with best practices reported in prior SER studies \cite{shendabadi}. All experiments are optimized using the AdamW optimizer. Models are trained for a maximum of 30 epochs with a peak learning rate of 0.001. A cosine learning rate scheduler is employed, along with a warm-up phase covering the first 10\% of the total training steps to stabilize optimization. The batch size is set to 16 for both training and validation.
\begin{table*}[t]
\centering
\caption{Results of Whisper-Small Variants in Our Experiments}
\label{tab:ablation}
\resizebox{\textwidth}{!}{%
\begin{tabular}{lccccc}
\toprule
\textbf{Experiment} & \textbf{WA (\%)} & \textbf{UA (\%)} & \textbf{F1 (\%)} & \textbf{Precision (\%)} & \textbf{Recall (\%)} \\
\midrule
Vanilla Whisper + FC Projector & $87.73 \pm 2.85$ & $80.52 \pm 3.24$ & $87.49 \pm 2.86$ & $87.87 \pm 2.83$ & $87.73 \pm 2.85$ \\
Vanilla Whisper + PCA & $89.50 \pm 2.29$ & $82.80 \pm 4.75$ & $89.35 \pm 2.31$ & $89.78 \pm 2.36$ & $89.50 \pm 2.29$ \\
Persian Finetuned Whisper + PCA & $\mathbf{89.72} \pm 1.89$ & $\mathbf{83.73} \pm 4.48$ & $\mathbf{89.61} \pm 1.91$ & $\mathbf{89.93} \pm 2.01$ & $\mathbf{89.72} \pm 1.89$ \\
\bottomrule
\end{tabular}%
}
\end{table*}
Model performance is evaluated using multiple metrics, including Weighted Accuracy (WA), Unweighted Accuracy (UA), F1-score, Precision, and Recall. Given the pronounced class imbalance in the ShEMO dataset, UA and F1-score are considered the primary evaluation criteria, as they provide a more reliable assessment of performance across all emotion categories.
\section{Results and Analysis}
The experimental results suggest that replacing the learned fully connected projection layer with PCA for dimensionality reduction of frame-level Whisper representations is associated with consistent performance gains under our experimental setup. In a controlled comparison where all other components are kept identical, the PCA-based approach yields improvements on the order of 2\% in Weighted Accuracy and 3\% in Unweighted Accuracy when reducing the 768-dimensional Whisper-small representations to 512 dimensions.

\begin{table*}[h!]
\centering
\caption{Performance Comparison with Prior Work on Persian SER}
\label{tab:comparison}
\begin{tabular}{lcc}
\toprule
\textbf{Base Model} & \textbf{WA (\%)} & \textbf{UA (\%)} \\
\midrule
Wav2Vec~2.0 Large~\cite{nasersharif}      & 86.80 & 80.60 \\
HuBERT Large~\cite{ma}             & 83.35 & 64.29 \\
WavLM Large~\cite{ma}               & 87.13 & 71.72 \\
Data2Vec~2.0 Large~\cite{ma}     & 82.68 & 64.09 \\
Whisper Large V3~\cite{ma}        & 89.55 & 80.23 \\
Whisper Small~\cite{shendabadi}           & 89.19 & 83.07 \\
\textbf{Whisper Small (Ours)}             & \textbf{89.72} & \textbf{83.73} \\
\bottomrule
\end{tabular}
\end{table*}

In addition to performance differences, PCA offers notable computational advantages. Specifically, it results in a substantial reduction in training latency---approximately 30\% per epoch in our experiments---and lowers memory usage by eliminating additional trainable parameters. These efficiency benefits may be particularly relevant in multimodal emotion recognition settings, where multiple high-dimensional feature extractors are commonly employed.

We also examine the impact of adapting Whisper to the target language through Persian ASR fine-tuning. While one might naturally hypothesize that language-specific fine-tuning would enhance downstream SER performance by improving linguistic representations, the observed gains are relatively modest. Across evaluation metrics, ASR fine-tuning yields improvements of less than 1\% in most cases. Similar observations have been reported in recent work on Spanish SER \cite{b3}, where a Spanish-fine-tuned Whisper large-v2 model showed limited improvement in some datasets and no improvement in others over the original pretrained model for emotion recognition.

Several plausible explanations may account for this behavior. One hypothesis is that, although linguistic and semantic information can contribute to emotion recognition, SER systems trained solely on speech signals may rely more heavily on prosodic and paralinguistic cues, thereby reducing the impact of enhanced linguistic representations. Another possible explanation is that ASR fine-tuning primarily benefits the decoder components of Whisper, which are critical for reducing word error rate, whereas downstream SER relies exclusively on encoder representations. As a result, the factors driving ASR improvements may not directly transfer to emotion recognition performance.

However, an additional analysis of parameter updates does not strongly support this latter hypothesis. Specifically, examining the absolute changes in the weights of the encoder and decoder layers during ASR fine-tuning reveals no clear difference in their overall update patterns, with both components exhibiting parameter changes of a similar order of magnitude. This observation reduces the likelihood that the limited transfer to downstream SER performance can be attributed solely to the dominance of decoder-side adaptations during ASR fine-tuning.

\section{Conclusion}
In this paper, we addressed key challenges in Persian SER under low-resource conditions by proposing an efficient framework based on Whisper encoder representations. We demonstrated that unsupervised dimensionality reduction using PCA not only eliminates a substantial number of trainable parameters but also consistently improves emotion recognition performance compared to a learned projection layer, while reducing training latency and memory usage. Furthermore, we systematically investigated the effect of language-specific ASR fine-tuning on downstream SER and found that the resulting gains are modest, suggesting that improved linguistic representations do not strongly transfer to emotion-related tasks under the evaluated conditions; the underlying reasons for this limited transfer remain an open question for further research. Notably, our proposed framework achieves state-of-the-art results on the ShEMO dataset, surpassing previously reported performance across both weighted and unweighted accuracy metrics.

\section*{Aknowledgment}
The authors acknowledge the use of GPT-5.4 to assist in manuscript preparation, language editing, and improving the clarity of the text. All AI-generated suggestions were reviewed and revised by the authors, who take full responsibility for the final content of this paper.

\section*{Code Availability}
The implementation and scripts required to reproduce the experiments are available at: \url{https://github.com/alishendabadi/PersianSERwithPCAandASRFinetune}.

\section*{Data Availability} 

The data that support the findings of this study are available from the following sources: the ShEMO dataset is accessible via its GitHub repository at \url{https://github.com/mansourehk/ShEMO?tab=readme-ov-file}, and the Persian Youtube dataset for ASR finetuning is accessible via its Huggingface repository at \url{https://huggingface.co/datasets/pourmand1376/asr-farsi-youtube-chunked-30-seconds}.

\section*{Author Contributions}

\noindent\textbf{Ali Shendabadi}: Conceptualization, Software, Validation, Methodology, Investigation, Data Curation, Writing Original Draft.

\noindent\textbf{Parnia Izadirad}: Software, Validation, Visualization, Review \& Editing.

\noindent\textbf{Mostafa Salehi}: Supervision, Review \& Editing, Project Administration.

\section*{Funding}

This research did not receive any specific grant from funding agencies in the public, commercial, or not-for-profit sectors.

\bibliography{sn-bibliography}

\end{document}